\documentclass[letterpaper,conference]{IEEEtran}
\usepackage{url}
\IEEEoverridecommandlockouts
\usepackage{float}

\usepackage{cite}
\usepackage{algorithm}
\usepackage{algorithmic}
\usepackage{amsmath,amssymb,amsfonts}
\usepackage{wrapfig}
\usepackage{graphicx}
\usepackage{textcomp}
\usepackage{xcolor}
\usepackage{graphicx}
\usepackage{hyperref}
\usepackage{float}

\usepackage{subcaption}  % preferred over subfig
\hypersetup{draft}   % disables link creation
\def\BibTeX{{\rm B\kern-.05em{\sc i\kern-.025em b}\kern-.08em
    T\kern-.1667em\lower.7ex\hbox{E}\kern-.125emX}}

\usepackage[letterpaper,
  left=0.65in,    % Increased from 0.62in for safety margin
  right=0.65in,   % Balanced with left margin
  top=0.8in,
  bottom=1.1in,   % Increased from 1in to ensure >1in minimum
  columnsep=0.24in,
  textheight=8.75in
]{geometry}
\begin{document}

\title{\fontsize{23pt}{25pt}\selectfont 
STAC: Spatio-Temporal Associations for Cross-Camera Analytics in Bandwidth-Constrained Environments
}

\author{
Ragini Gupta$^{*}$, Lingzhi Zhao$^{*}$, Jiaxi Li$^{*}$, Volodymyr Vakhniuk$^{*}$, Claudiu Danilov$^{+}$,  Josh Eckhardt$^{+}$, \\
Keyshla Bernard$^{+}$, Klara Nahrstedt$^{*}$\\
$^{*}$University of Illinois Urbana-Champaign, Urbana, IL, USA\\
$^{+}$Boeing Research and Technology, USA\\
\{raginig2, lz26, jiaxili3, vv9, klara\}@illinois.edu, \{cdanilov, josh.d.eckhardt, keyshla.j.bernard\}@boeing.com
}
\maketitle
\vspace{-0.8cm}
\begin{abstract}
In IoT based distributed network of cameras, real-time multi-camera video analytics is challenged by high bandwidth demands and redundant visual data, creating a fundamental tension where reducing data saves network overhead but can degrade model performance, and vice versa. We present STAC, a cross-cameras surveillance system that leverages spatio-temporal associations for efficient object tracking under constrained network conditions. STAC integrates multi-resolution feature learning, ensuring robustness under variable networked system level optimizations such as frame filtering, FFmpeg-based compression, and Region-of-Interest (RoI) masking, to eliminate redundant content across distributed video streams while preserving downstream model accuracy for object identification and tracking. Evaluated on NVIDIA's AICity Challenge dataset, STAC achieves a 76\% improvement in tracking accuracy and an 8.6× reduction in inference latency over a standard multi-object multi-camera tracking baseline (using YOLOv4 and DeepSORT). Furthermore, 29\% of redundant frames are filtered, significantly reducing data volume without compromising inference quality. 
\end{abstract}
\vspace{-0.4cm}
\section{Motivation and Challenges}
\vspace{-0.12cm}
Large-scale Internet-of-Things (IoT) networks of distributed cameras are increasingly deployed in diverse environments to support applications such as traffic monitoring \cite{crossroi}, retail stores \cite{retailstores2}, criminal tracking, forensic investigation, and public surveillance \cite{milcom100}. Many incorporate overlapping Fields of View or FoVs (e.g., road intersections, airports, national borders), enabling richer analysis by capturing details missed in one view such as an occluded person to be captured from another angle. In security and defense contexts, distributed cameras are also deployed as standalone units \cite{milcom100,milcom200} or on drones/UAVs \cite{milcom300} to enhance situational awareness by identifying enemy positions and monitoring troop movements, thereby enabling coordinated decision-making. These networks are often organized in a three-tier architecture (e.g., Fig.~1 illustrates the NVIDIA AICity dataset \cite{aicity} from a multi-camera people-tracking setup at a retail-store capturing different overlapping views), where video streams are sent to an edge server for initial processing, and the processed data or extracted features are then forwarded to the cloud for further analytics and query-based retrieval. While on-camera analytics can reduce reliance on external infrastructure, limited compute and memory resources often constrain their capabilities, leading many deployments to offload to nearby edge servers or cloud. For analytics, DNN-based object tracking methods such as YOLO~\cite{crossroi} and FastCNN~\cite{ren2016faster} are used to detect and track objects across different times and locations from cameras. However, real-time inference on multiple camera videos demands high network bandwidth for communication and introduces latency and reliability issues, conditions that are ill-suited for tactical networks with constrained resources, unstable links, packet loss and frequent disconnections. While advanced video encoding (e.g., High Efficiency Video Coding such as H.264 \cite{hevc}) and filtering techniques \cite{highaccuracy2} can reduce bandwidth usage, high-quality tracking models remain sensitive to compression artifacts and frame drops, leading to accuracy degradation \cite{multiconstrained,reid2,argus}. This happens because models accuracy is tightly coupled to system-level variable knobs such as frame bitrate, video resolution, or codec scheme and are not robust to changes in these settings, further limiting their adaptability in resource-constrained networks. On the other hand, many systems treat each camera as a passive streamer, analyzing feeds independently \cite{chameleon, nsdi} without leveraging proximity-based collaboration or exploiting spatio-temporal redundancies, leading to unnecessary computation.
\begin{figure}[!t]
\vspace{-0.2cm}
    \centering
    \vspace{-0.23cm}
\includegraphics[width=0.48\textwidth]{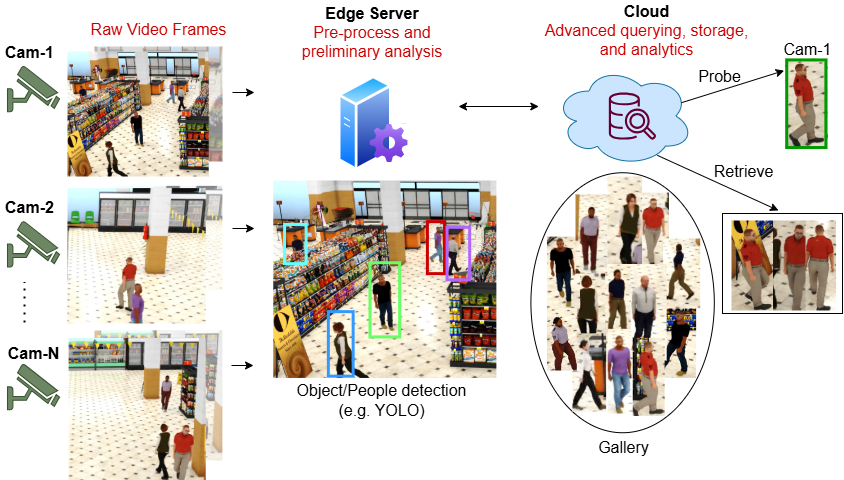}
      \vspace{-0.2cm}
    \caption{\small Three-tier architecture for a distributed camera network}
    
    \vspace{-0.25cm}
\end{figure}

To address these challenges, we present \textbf{\textit{STAC}}, an end-to-end, modular, cross-camera analytics framework for resource-constrained environments. STAC adopts an adaptive hierarchical architecture comprising \textit{cameras}, an \textit{edge server}, and the \textit{cloud}. Cameras perform local detection, apply the classic video encoder H.264 for compression using FFmpeg\cite{ffmpeg2}, and selectively offloads only unique informative frames, identified through contextual filtering, to the edge server. At the edge, STAC performs feature extraction, cross-camera object identity matching, and Region-of-Interest (RoI) masking sending data to the cloud, that is solely responsible for global coordination and long-term identity management. At its core, STAC targets the fundamental task of \textbf{multi-camera, multi-target tracking}, which requires locating and maintaining consistent identities for objects (i.e. people) across space (multiple cameras) and time (tracking within each camera over time). A key bottleneck in such deployments is the repeated execution of costly identification inference for the same individual on multiple cameras. Moreover, accurate association and identity maintenance for people in distributed camera network remains challenging due to (a) appearance changes (e.g., clothing), (b) pose and viewpoint variations, (c) low frame quality from compression or poor lighting, (d) inconsistent scaling of person features, and (e) the absence of strong temporal or contextual cues in crowded or occluded scenes. Existing multi-camera tracking systems often optimize association between cameras at a coarse-grained level, such as groups of cameras \cite{rel1} or individual camera-wise filtering \cite{spatula,rel2,argus}. While these approaches reduce computation, their benefits are marginal in dense deployments with overlapping fields of view. For e.g., camera-level association \cite{spatula, rel2} fails to exploit fine-grained redundancies when the same target appears across multiple cameras simultaneously. Similarly, group-level methods \cite{rel1} prioritize scalability over precision, trading identity consistency in crowded or overlapping scenes. In contrast, STAC targets fine-grained, per-object association by leveraging omni-scale features and cross-camera fusion. Specifically, STAC maintains spatio-temporal associations of objects, capturing their bounding boxes and identities across multiple cameras and over time within each camera stream. It then matches these associations to ensure identity consistency, avoiding repeated identification inference and appearance-based matching whenever location and temporal cues provide sufficient confidence. Omni-scale \cite{osnet} feature learning allows STAC to capture both global attributes (e.g., body size, clothing color) and fine-grained details (e.g., accessories), enabling robust associations even under appearance, pose, or lighting variations. This robustness ensures that system-level optimizations—including filtering, compression, and RoI masking—reduce computation and transmission costs without degrading tracking accuracy. It is worth mentioning that STAC’s modular design allows its OSNet feature extraction module to integrate with diverse multi-camera tracking models, enabling independent tuning of filtering and compression without impacting tracking accuracy.

\textbf{Contributions.} Key contributions of this work include: \textbf{(a)} an intelligent camera-edge-cloud framework that reduces network overhead via combination of optimization techniques like compression, selective offloading, and RoI masking while maintaining downstream model accuracy, \textbf{(b)} robust cross-camera tracking using omni-scale features and fine-grained spatio-temporal association between objects of interest to handle appearance changes and compression artifacts, and \textbf{(c)} a modular design that decouples system optimizations from tracking accuracy, achieving 4.8× bandwidth reduction and 76\% higher accuracy than the standard baseline model for multi-camera tracking (with a fixed scale CNN) on the AICity NVIDIA multi-camera dataset.
\begin{figure}[t]
    \centering
\includegraphics[width=0.48\textwidth]{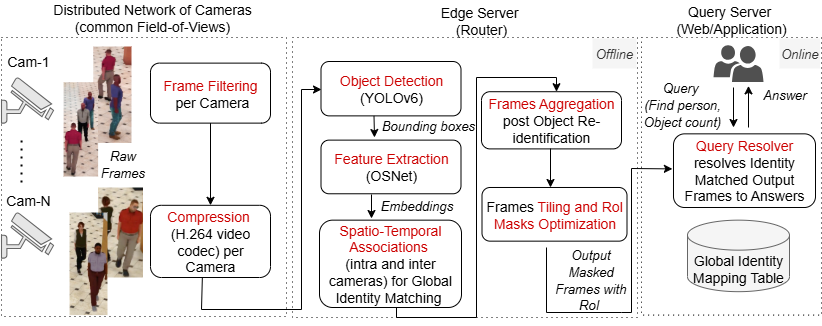}
    \caption{\small System workflow}
    \label{fig:personreid}
\end{figure}
\vspace{-0.2cm}
\section{System Design}
\vspace{-0.2cm}
The system has two main goals: (a) identifying common regions-of-interest (ROIs- with people) across cameras, and (b) streaming cross-camera correlations efficiently without compromising inference accuracy. Based on these, STAC is designed in two stages: an offline phase for model training and metadata preparation, and an online phase for real-time query resolution. Fig. 2 represents the STAC's system workflow. In the offline stage,  camera feeds are first forwarded to an intermediate edge router where frame filtering and H.264 compression is executed to eliminate redundancy. YOLOv6 is then used for person detection, followed by OSNet for identity feature extraction and spatio-temporal association across cameras by assigning global people IDs across cameras (a process known as object re-identification, or \textbf{\textit{ ReID \cite{crossroi}}}). ReID in this context involves three steps: (i) detecting object bounding boxes in each camera frame, (ii) extracting discriminative appearance features for each detection and comparing them to query features using similarity measures (e.g., cosine similarity), and (iii) aggregating identity matches across cameras to produce tracklets for downstream application logic (hereinafter, we refer to spatio-temporal associations between cameras as ReID for rest of the paper).
The server then aggregates the ReID outputs, masks each frame to retain only ROIs, and transmits these tiles with their metadata (ID, timestamp, embeddings) to query server. In the online phase at the query server, responses to user queries (e.g., person search, counting) are generated using the matched identity metadata. We assume a network of \texttt{N} cameras with their respective frame rate and duration.
\vspace{-0.3cm}
\section{System Components}
\vspace{-0.24cm}
\subsection{Offline phase}
The offline phase occurs prior to any live query activity and is executed entirely at the edge server, which has access to video feeds from all $N$ cameras. This phase includes multi-camera data acquisition, frame pre-processing (filtering and compression), person detection, feature embedding, cross-camera identity association, and model training for the ReID backbone. Key components are outlined below.
\vspace{-0.1cm}
\subsubsection{Video data collection}
The system simulates $N$ video streams ${C_1, \dots, C_N}$, each monitoring a shared environment. At each time step $t$, one frame is captured from each camera, forming a spatial batch $F_t = {f_t^1, f_t^2, \dots, f_t^N}$, where $f_t^i$ is the frame from camera $C_i$. Over $T$ steps, each camera $C_i$ produces a temporal sequence $T_i = {f_1^i, f_2^i, \dots, f_T^i}$. The person ReID system learns a mapping $\phi$ to associate individuals across space and time. Given a query person $q$ in frame $f_{t_0}^i$, the goal is to retrieve all frames $f_{t'}^j$ (for $j \in {1,\dots,N}$ and $t' \in {1,\dots,T}$) such that $\phi(q, f_{t'}^j) = \text{True}$ if $f_{t'}^j$ contains the same person.
\vspace{-0.1cm}
\subsubsection{Frame filtering and person detection using YoLov6} Once a batch of $N$ spatially-aligned frames is collected, object detection is applied to identify human subjects. We use YOLOv6 \cite{yolov6}, a real-time detector with strong accuracy-speed tradeoff, suitable for multi-camera systems. YOLOv6 runs independently on each frame and returns bounding boxes, class labels, and confidence scores.We keep only \texttt{person}-labeled detections with confidence scores above a set value \footnote{Detection confidence: $C = 0.45$ }. The threshold is empirically chosen to balance precision and recall in crowded scenes, ensuring that most true positive person instances are retained. The detected regions are cropped from frames and sent to the feature extractor for embedding and identity association.
\vspace{-0.1cm}
\subsubsection{Feature Extraction using OSNet}Omni-Scale feature learning Network (OSNet) \cite{osnet} is a deep CNN model that includes a residual block composed of multiple convolutional streams, each capturing features at a specific scale.  It learns both homogeneous features (e.g., body size, clothing color) and heterogeneous features (e.g., fine-grained details like bags or accessories) by fusing multiple spatial scales. As shown in Fig. 3, it uses a specialized residual block composed of parallel depth-wise separable convolutional layers, each capturing features at different receptive fields. Streams with smaller receptive fields (R=3) capture fine-grained, homogenous features, while larger receptive fields (R=7) capture high-scale features. To maintain efficiency, OSNet employs Lite 3x3 convolutions (1x1 convolutions followed by group-wise 3x3 convolutions), reducing computation and parameter count. A channel-wise aggregation gate (a mini-network) fuses these features into compact, discriminative embeddings. This suits our system's resource constraints with efficient identity extraction. Cropped person detections are passed through OSNet~\cite{osnet}. OSNet outputs a 512-dimensional embedding per detection, capturing multi-resolution features. For each of the $N$ frames, we obtain $M$ embeddings from the detected persons. These features enable identity matching across frames and cameras.
\vspace{-0.1cm}
\subsubsection{Identity association using similarity measurement} Our Re-ID system links identities across time (intra-camera) and space (inter-camera). We use OSNet embeddings trained with triplet and cross-entropy losses to cluster same-person features and separate different identities. Temporal association (intra-camera) is done by comparing embeddings between consecutive frames of the same camera. A similarity matrix $S \in \mathbb{R}^{N \times M}$ is computed as $S = 1 - D$, where $D$ is the pairwise Euclidean distance. The current detection (i.e. bounding box) is assigned the identity of the most similar detection from the previous frame if the embedding similarity between them exceeds a set threshold\footnote{Temporal association threshold $\tau'_{\text{sim}} = 0.65$.}. Spatial association (inter-cameras) is handled by computing cosine similarity between embeddings across all $N \times N$ camera pairs. For each pair $(C_i, C_j)$, embeddings are compared, and a match is declared if the similarity exceeds a preset spatial association threshold\footnote{Spatial association threshold: $\tau_{\text{sim}} = 0.7$.}. When a match is found, we assign a consistent identity label to detections (bounding boxes) with similar embeddings across different camera views. This enables a consistent global identity assignment for each person across different cameras, relying solely on semantic similarity in the learned embeddings space. The use of nearest-neighbor matching is made effective due to the discriminative power of OSNet's embeddings, which are trained with triplet and cross-entropy loss to ensure high inter-class separation and low intra-class variance \cite{osnet}. These embeddings enable reliable identity matching across cameras.
\vspace{-0.35cm}
\begin{figure}[h] % r = right side, width reserved for figure
\vspace{-0.2cm}
    \centering
    \includegraphics[width=0.65\linewidth]{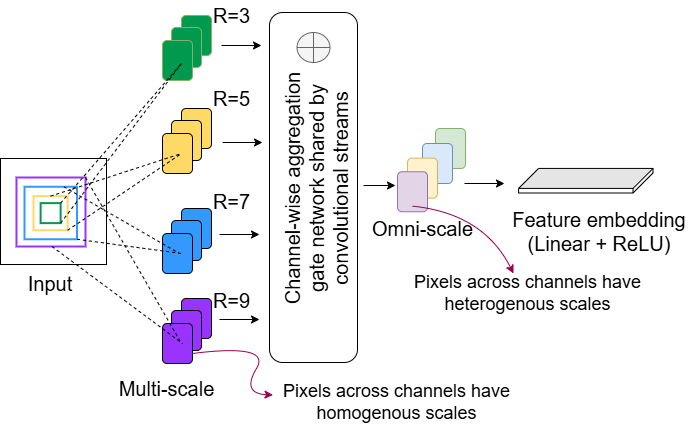}
    \caption{\small OSNet architecture (R is receptive field size) \cite{osnet}
}
    \vspace{-0.2cm}
\end{figure}
\par
\vspace{-0.4cm}
\subsection{Bandwidth optimization}
\vspace{-0.1cm}
To operate efficiently in resource-constrained settings, our system uses bandwidth optimizations that reduce redundant visual data while preserving inference quality. These are applied both before and after ReID. Pre-ReID, frames are filtered and compressed to remove spatial redundancy. Post-ReID, tile-based selection transmits only regions with detected persons. This ensures scalable, accurate performance across multi-camera deployments.
\subsubsection{Pre ReID - Frame filtering and compression}
To reduce visual redundancy before ReID, we apply a lightweight frame filtering and compression module. The goal is to remove near-duplicate frames across camera streams while preserving those critical for identity association. Filtering uses perceptual (SSIM) and pixel-level (MSE) similarity metrics to detect redundant content. Frames with high similarity and low difference are dropped, using conservative thresholds to retain meaningful changes such as new person entries or viewpoint shifts. It is important to note that this filtering step is designed to suppress background or static content duplication, not to eliminate different views of the same person. Since ReID relies on diverse visual perspectives, we preserve all frames that may contribute to person identity matching. 
After filtering, the retained frames are compressed using H.264 encoding via \texttt{ffmpeg}'s \texttt{libx264} codec~\cite{ffmpeg2}, which reduces bitrate and resolution while maintaining visual quality. The output is a reduced and compressed set of frames, which is then passed to the ReID module for further processing.
\vspace{-0.1cm}
\subsubsection{Post ReID - Frames Tiling and RoI Masks Optimization} Following person re-identification we apply a masking procedure to reduce transmission overhead by preserving only regions of interest (with people detections) from different cameras' frames. Each frame is divided into rectangular tiles in a grid (e.g., 6×4), then solves an optimization to select the smallest tile set covering all object appearances across cameras. Adjacent tiles are merged for efficiency, and non-RoI regions are discarded during transmission (i.e. tiles that do not intersect with any bounding box of query-matched detections are masked out), see Algorithm-1. This process operates on the output of the ReID system, where detections have already been assigned consistent global IDs using OSNet embeddings. At this stage, no further embedding comparisons are needed, only spatial processing of labeled detections. The masking logic ensures only tiles with valid matches are retained, removing both empty areas and redundant appearances in overlapping views, thus, minimizing data transmission. STAC adapts spatial optimization principles from CrossRoI \cite{crossroi}, originally developed for vehicle ReID, adapted to our multi-camera person tracking and identification with overlapping FoVs.  
\vspace{-0.2cm}
\subsection{Online inference} \vspace{-0.14cm}The online phase handles real-time inference and query resolution from live video streams across $N$ cameras. Each frame is processed using YOLOv6 for person detection, followed by OSNet for identity embedding and cross-camera association. This yields consistent person IDs, bounding boxes, and appearance metadata. Queries like “how many times did Person P1 appear?” are answered using stored IDs, timestamps, and features. When visual evidence is required, the system transmits only relevant spatial tiles instead of full frames, significantly reducing bandwidth. These tiles retain regions of interest while masking irrelevant areas, enabling low-latency, bandwidth-efficient, and semantically meaningful responses in real-time.
In the current implementation, we evaluate the system's capability using simple identity lookup based queries, for example, retrieving all correctly identified person instances across cameras. This framework lays the foundation for supporting more advanced query types in future deployments. These include \textit{counting queries} (e.g., ``How many distinct persons appeared?''), \textit{attribute-based queries} (e.g., ``Who wore color $C$?''), and \textit{temporal queries} (e.g., ``When did person $P_i$ enter?''), all enabled by the same metadata: consistent person IDs, timestamps, and visual appearance features, while tile-based regions-of-interest transmission per frame maintains bandwidth efficiency for all query types. 
\begin{algorithm}[!t]
\vspace{-0.1cm}
\small
\caption{\small Frame Tiling and RoI Masks Optimization}
\label{alg:tiling-masking}

\begin{algorithmic}
\STATE \hspace*{-0.2cm} \textbf{Input:} For each Camera $i$: video frames $\{F_{i,t}\}$; object bounding boxes with IDs $\mathcal{B}_{i,t}$ from ReID stage; tile grid $\mathcal{G}_i$.
\STATE \hspace*{-0.2cm} \textbf{Output:} Optimized RoI mask $M_i$ per Camera.
\STATE \hspace*{-0.2cm} \textbf{Tile partitioning:} Divide each frame $F_{i,t}$ into a fixed grid of tiles $\mathcal{G}_i = \{\tau_1, \tau_2, \dots\}$.
\STATE \hspace*{-0.2cm} \textbf{Per-frame RoI tiles:} Per frame at $t$, mark tiles that overlap with any bounding box in $\mathcal{B}_{i,t}$ (objects of interest by ID).
\STATE \hspace*{-0.2cm} \textbf{Aggregate coverage:} Form coverage matrix $A$ where $A_{o,\tau} = 1$ if tile $\tau$ covers object $o$ in any frame.
\STATE \hspace*{-0.2cm} \textbf{Optimization:} Solve
\STATE \hspace*{-0.2cm} \(\displaystyle \min_{z \in \{0,1\}^{|\mathcal{G}_i|}} \sum_{\tau} z_\tau
         \text{     s.t. } \forall o\,  \sum_{\tau} A_{o,\tau}\, z_\tau \ge 1 \)
\STATE where \(z_\tau = 1\) indicates tile \(\tau\) is selected, to find smallest tile set \(M_i \subseteq \mathcal{G}_i\) covering all object appearances.
\STATE \hspace*{-0.2cm} \textbf{Tile grouping:} Merge adjacent tiles in $M_i$ into larger rectangles to improve compression efficiency.
\STATE \hspace*{-0.2cm} \textbf{Online masking:} For each frame $F_{i,t}$, black out all pixels outside $M_i$; encode and transmit only RoI tiles $\tau \in M_i$.
\end{algorithmic}
\vspace{-0.1cm}
\end{algorithm}

\section{Experimental Setup}
\vspace{-0.27cm}
\subsection{Datasets}
\vspace{-0.1cm}
We evaluate STAC on the well-known AICity Challenge 2023 Multi-Camera People Tracking dataset published by NVIDIA \cite{aicity}. The dataset, collected using the NVIDIA's Omniverse platform, consists of high-resolution 1080p, 30 FPS indoor videos (11–12 minutes each) from multiple cameras at different positions with overlapping views of a retail store. 
\vspace{-0.16cm}
\subsection{Baseline model}
\vspace{-0.1cm}
We benchmark our system against a widely adopted multi-camera tracking baseline that combines YOLOv4 for detection and DeepSORT for tracking \cite{deepsort}. This baseline represents a standard two-stage approach in which YOLOv4 first detects people in each frame, after which DeepSORT associates these detections across frames using motion and appearance cues. The motion component employs a Kalman filter to predict object states during occlusions, while the appearance model uses CNN embeddings trained with weighted softmax and triplet loss for discriminative feature learning. Matching occurs through Euclidean distance comparisons between embeddings, enabling identity persistence across frames.
\vspace{-0.16cm}
\subsection{Evaluation metrics}
\vspace{-0.1cm}
Evaluation of multi-object multi-camera tracking systems typically consider three aspects: detection (ability to identify objects in a frame), localization (ability to accurately locate them), and association (ability to determine whether objects across frames are the same or different). A person is considered correctly identified if: (1) the predicted bounding box sufficiently overlaps with the ground truth (Intersection-over-Union, IoU $\geq \tau_{sim}$) and (2) the assigned ID matches the ground truth label. An identification is incorrect if either criterion fails due to a missed/poorly localized detection (detection/localization error) or an incorrect cross-camera ID assignment (association error). Subsequently, we adopt the well-known \textit{Mean Target Tracking Accuracy (MTTA)} metric from CLEAR MOT~\cite{mtta2,mtta}, expressed in percentage form, evaluates the proportion of correctly tracked objects relative to the total number of ground-truth objects across all frames and cameras, formulated as:
\vspace{-0.24cm}
\begin{equation}
\vspace{-0.13cm}
\text{MTTA} = \frac{1}{N} \sum_{n=1}^{N} \left( \frac{C_{\text{correct}}^{(n)}}{T_{\text{total}}^{(n)}} \right) \times 100\%
\vspace{-0.2cm}
\end{equation}
where $N$ is the total number of cameras, $C_{\text{correct}}^{(n)}$ is the number of correctly identified persons in camera $n$ (requiring both IoU $\geq \tau_{sim}$ and correct ID)\footnote{$\tau_{sim}=0.7$, ensures predicted boxes are well aligned with ground truth while allowing minor localization variance.}, and $T_{\text{total}}^{(n)}$ is the total number of ground-truth persons in camera $n$. Additionally, we evaluate STAC’s system performance by measuring network overhead (bitrate reduction and bandwidth utilization) under varying uplink bandwidths, and the end-to-end respond latency.
\vspace{-0.2cm}
\section{Results} \vspace{-0.12cm}
\subsection{End-to-End (E2E) inference latency} \vspace{-0.1cm} We evaluate the per-frame end-to-end inference time of our pipeline against a baseline model using two camera streams from the AIChallenge dataset, trimmed to 13 seconds (approx. 716 frames) including portions of the scene that were more densely populated. As shown in Fig. 4(a), the baseline system (YOLOv4 + DeepSORT) takes an average of 14.82 seconds per frame, totaling 10,605 seconds for the full sequence. In contrast, our optimized pipeline processes each frame in 1.73 seconds on average, with a total time of 1,239 seconds. This yields an 8.6$\times$ latency improvement by reducing redundant frames via filtering, compression, and tile selection, and by using YOLOv6 for faster detection than baseline's YOLOv4.
\vspace{-0.3cm}
\vspace{-0.3cm}
\begin{figure*}[t]
    \centering
    % Latency first
    \begin{subfigure}[b]{0.15\textwidth}
        \includegraphics[width=\linewidth,height=2.8cm]{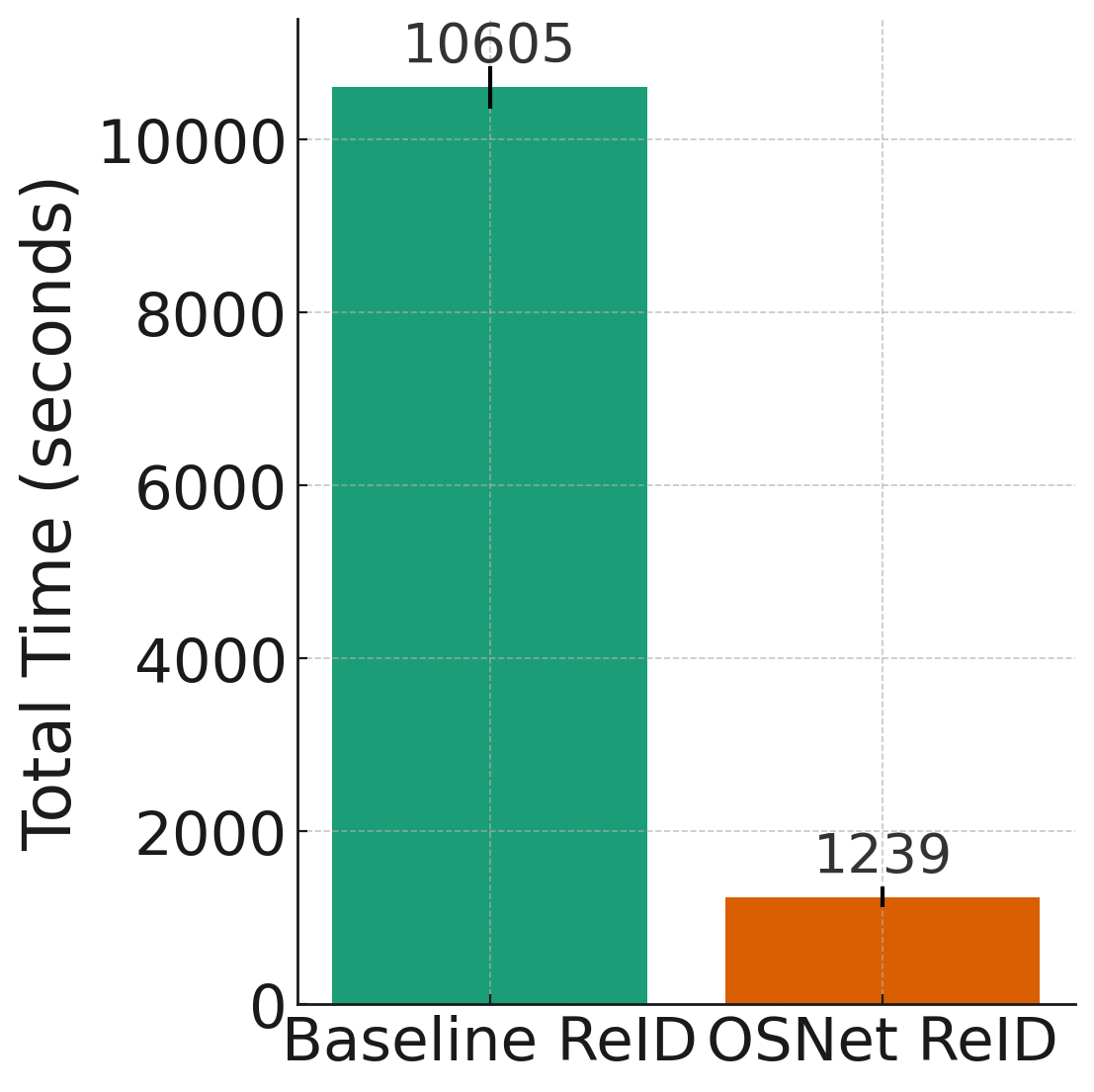}
        \vspace{-0.3cm}
        \caption{\scriptsize Latency}
        \label{fig:latency}
    \end{subfigure}
    \hspace{-0.2em}
    % Accuracy next
    \begin{subfigure}[b]{0.15\textwidth}
        \includegraphics[width=\linewidth,height=2.8cm]{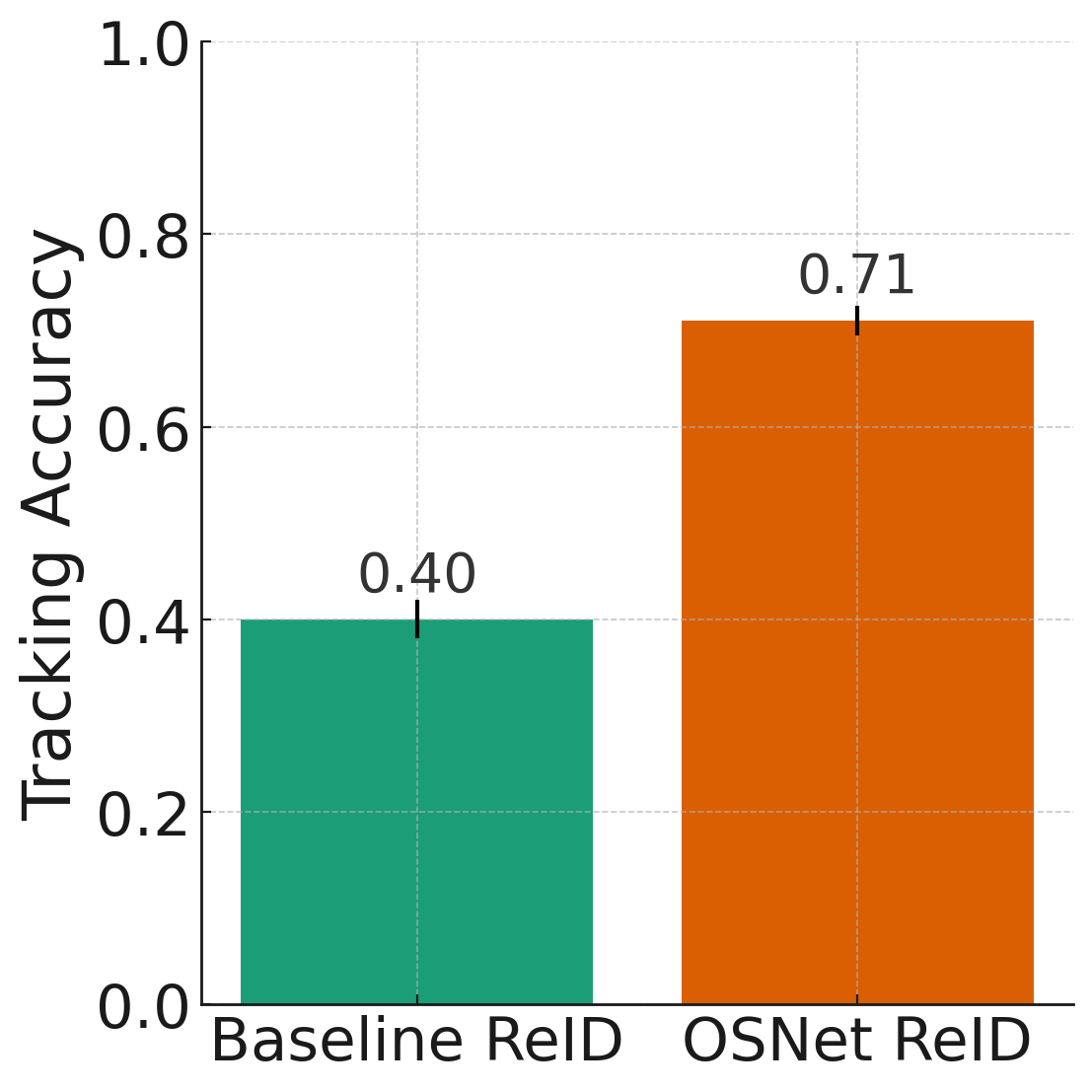}
        \vspace{-0.3cm}
        \caption{\scriptsize Accuracy}
        \label{fig:accuracy}
    \end{subfigure}
    \hspace{-0.2em}
    % SSIM vs. MSE filtering
    \begin{subfigure}[b]{0.35\textwidth}
        \includegraphics[width=\linewidth,height=4cm]{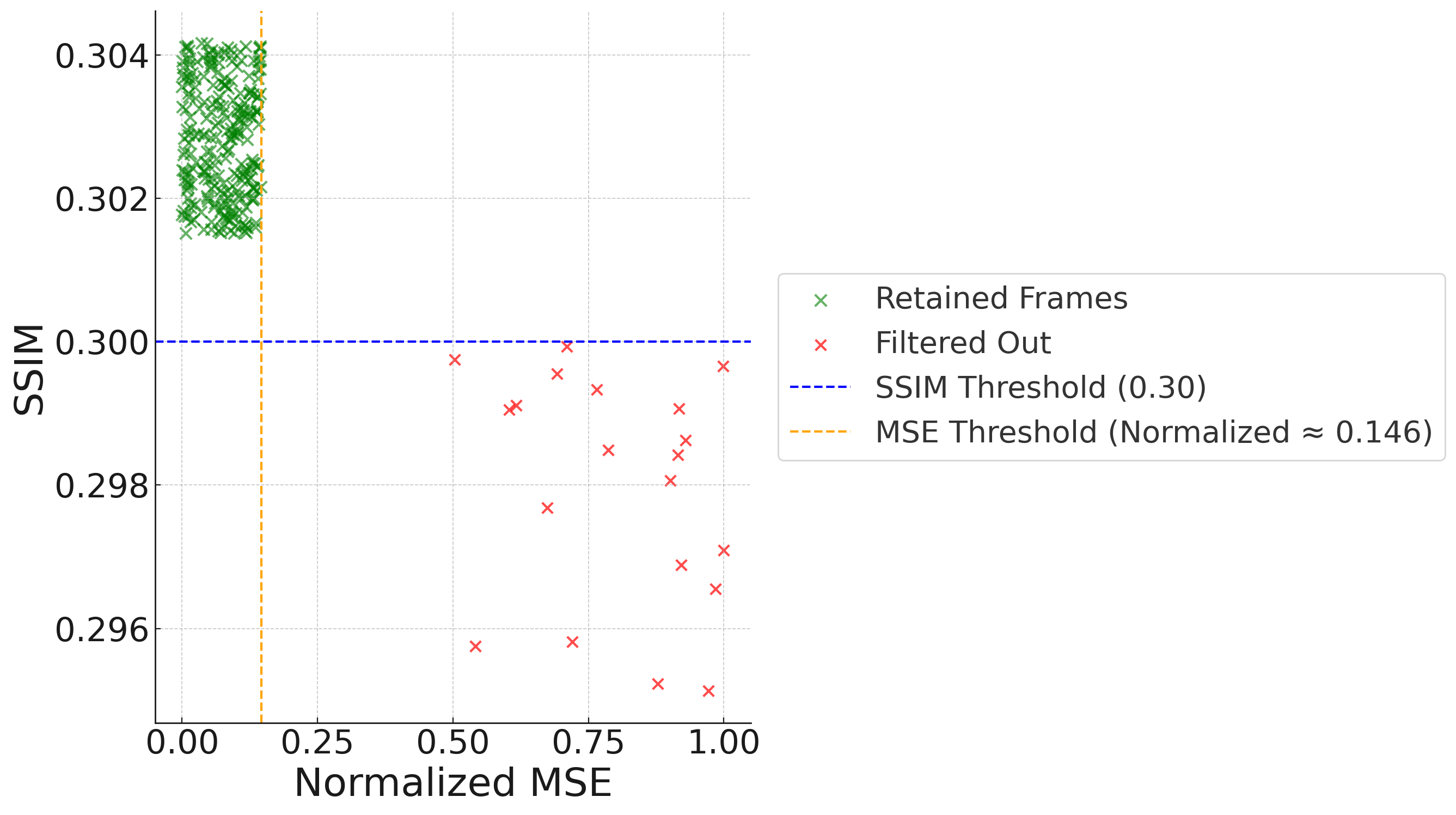}
        \caption{SSIM vs. MSE filtering}
        \label{fig:scatter}
    \end{subfigure}
    \hspace{-0.2em}
    % File size
    \begin{subfigure}[b]{0.15\textwidth}
        \includegraphics[width=\linewidth,height=2.8cm]{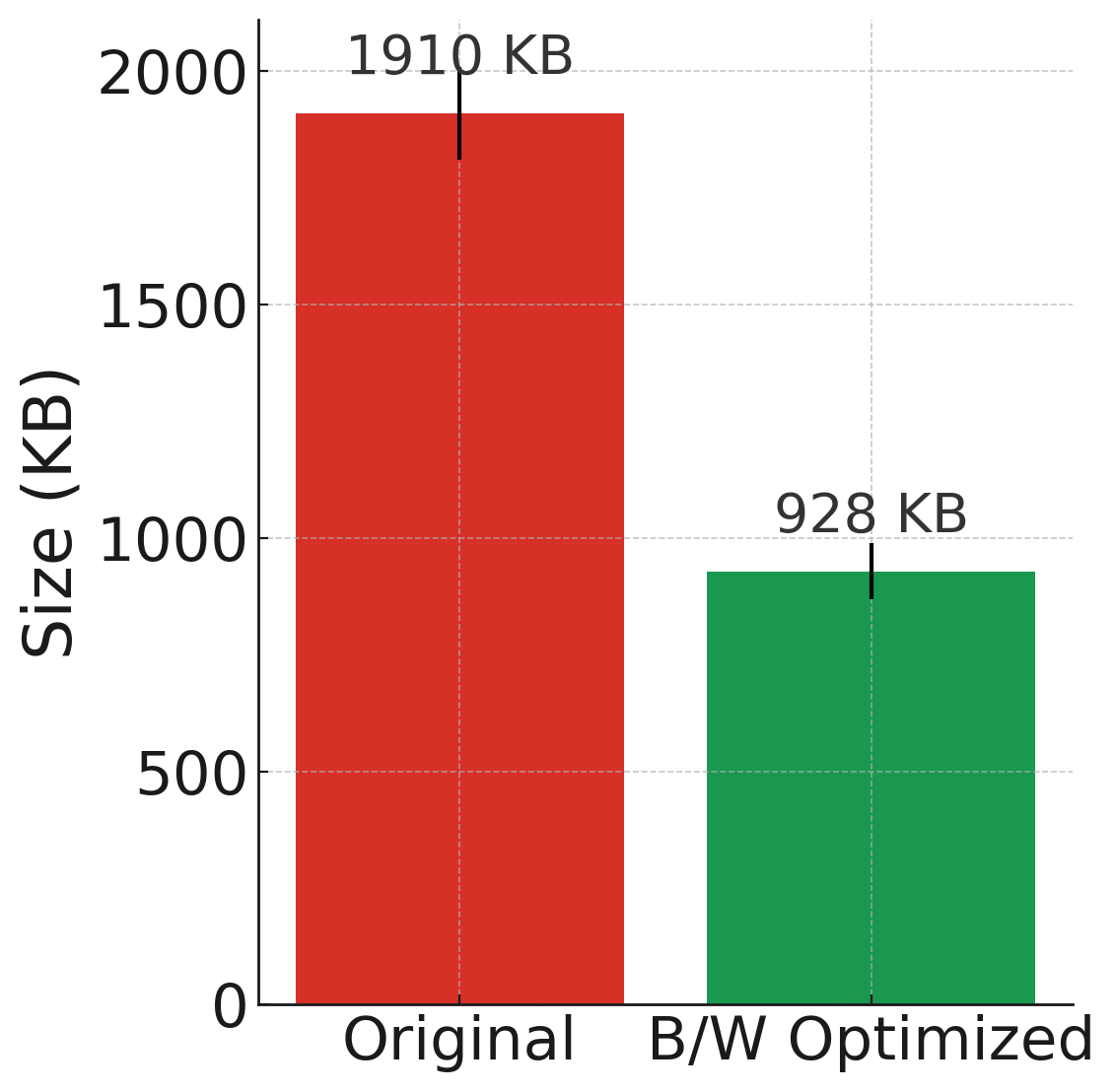}
        \vspace{-0.3cm}
        \caption{\scriptsize File size}
        \label{fig:size}
    \end{subfigure}
    \hspace{-0.2em}
    % Bitrate
    \begin{subfigure}[b]{0.15\textwidth}
        \includegraphics[width=\linewidth,height=2.8cm]{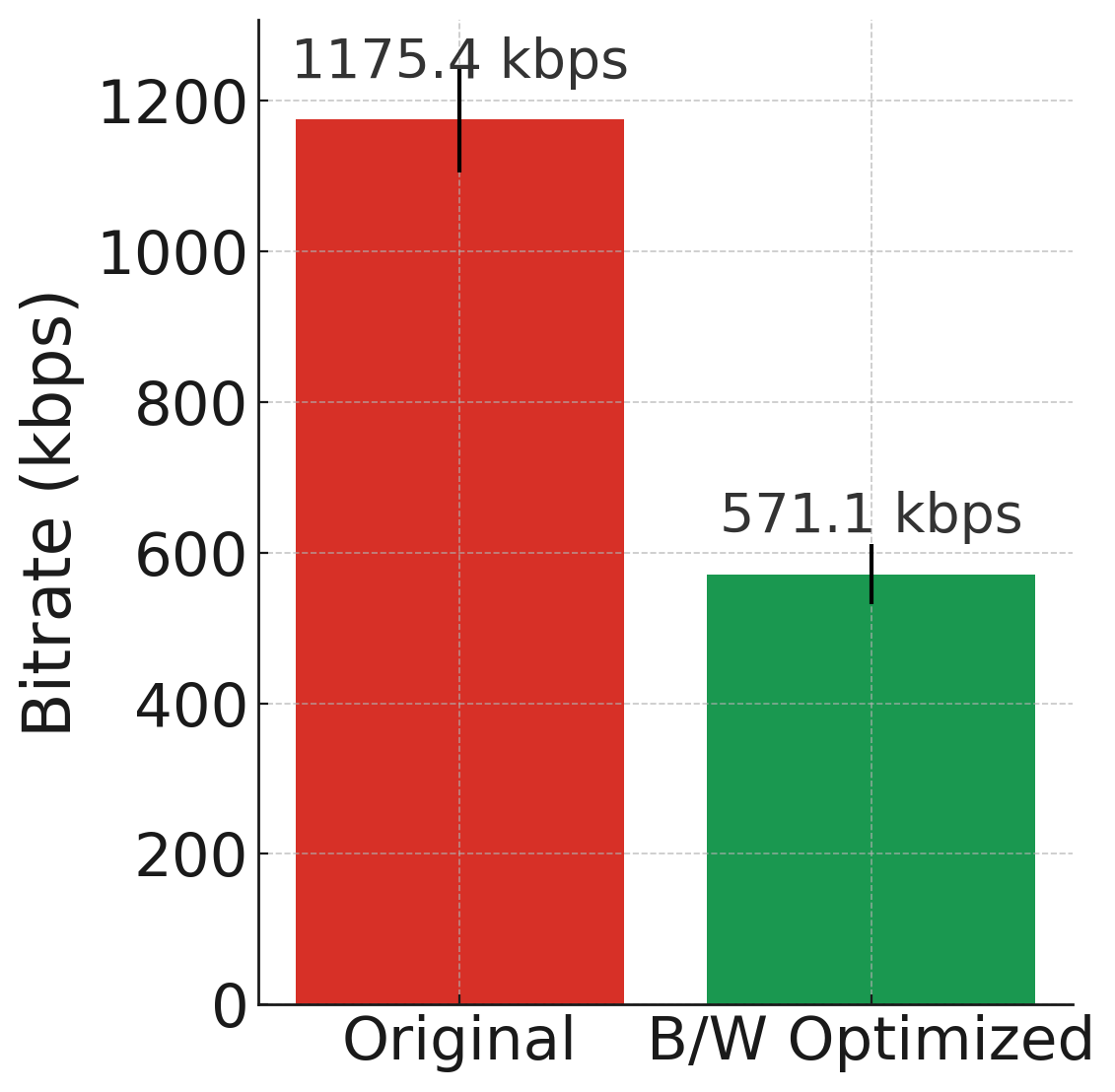}
        \vspace{-0.3cm}
        \caption{\scriptsize Bitrate}
        \label{fig:bitrate}
    \end{subfigure}

    \vspace{-0.2cm}
    \caption{\small STAC performance analysis showing (a) end-to-end latency, (b) tracking accuracy, (c) frame filtering analysis (SSIM vs. MSE), (d) file size reduction, and (e) bitrate optimization.}
    \label{fig:final_results}
\end{figure*}
\subsection{Tracking accuracy} \vspace{-0.15cm}
As shown in Fig. 4(b), the baseline model achieved a tracking accuracy of 0.40, while our STAC implementation with OSNet ReID model attained a significantly higher score of 0.707 for tracking individuals across cameras—a 76.8\% improvement. This difference is largely attributable to improved identity association across frames and cameras. The baseline tracking model is sensitive to frequent ID switches, leading to poor identity consistency even on raw video streams. On the other hand, STAC maintains high tracking accuracy even on filtered and compressed multi-camera streams. This is likely due to the use of ResNet-50 embeddings in the baseline, which are less discriminative for person re-identification. In contrast, our method exhibits more reliable identity mapping at different feature resolutions. Even when detection is sparse or when individuals exit and re-enter the frame, identity continuity is largely preserved with STAC. The accuracy gains are further reinforced by filtering redundant data within and across cameras, enabling the model to train on cleaner, unique frames, improving generalization and preventing overfitting. This robustness explains the substantial improvement in tracking accuracy alongside low end-to-end network overhead. Integrating STAC’s feature extraction into baseline tracking models is a promising future direction for enhancing robustness under bandwidth-constrained video streams.
\vspace{-0.23cm}
\subsection{Frame filtering analysis} 
\vspace{-0.2cm}
Our system implements a selective frame retention strategy using complementary structural similarity (SSIM) and mean squared error (MSE) criteria to optimize the tradeoff between data reduction and tracking accuracy. This dual-threshold approach preserves frames containing meaningful visual variations while systematically eliminating both redundant duplicates and excessively noisy frames that could compromise downstream processing. Fig. 4 (c) presents a scatter plot of frame pairs analyzed using SSIM and normalized MSE metrics, with distinct threshold boundaries demarcating the filtering criteria.  structurally valuable frames suitable for reliable re-identification and those that either introduce redundancy or potential matching errors. Removing dissimilar frames reduces false associations in re-identification, where system might incorrectly mismatch identities due to abrupt scene changes or structural inconsistencies. On the other hand, removing redundant frames where both cameras capture near-identical structural content reduces processing overhead by eliminating computations on visually repetitive data. By removing approximately 29\% of total input frames between cameras through this process, we achieve significant computational savings. The specific threshold values\footnote{Frame pairs are retained for SSIM $\geq$ 0.30 and normalized MSE $\leq$ 0.146.}were carefully determined through empirical validation to balance data reduction against preservation of critical identity-matching information. 
 
\vspace{-0.24cm}
\subsection{Network efficiency} 
\vspace{-0.12cm}
To evaluate the impact of our preprocessing steps on transmission efficiency, we measure both the file size and resulting bitrate of the video stream before and after applying our frame filtering and compression pipeline.
Fig. 4(d) illustrates the reduction in video file size after applying filtering and compression \texttt{ffmpeg} using the \texttt{libx264} codec. The original video, recorded at 1080p and 30 FPS over a 13-second duration, is reduced from 1910 KB to 928 KB—representing a 51.4\% size reduction. Correspondingly, as shown in Fig. 4(e), the effective frame bitrate decreases from 1175.4 kbps to 571.1 kbps. This significant drop is directly attributable to the elimination of redundant frames and the use of bandwidth-efficient video encoding. Following prior work on edge video analytics including NVIDIA’s DeepStream \cite{Nvidia} and Reducto frameworks \cite{highaccuracy2} which operate under uplink capacities ranging from 3 to 10 Mbps, we assume a 5 Mbps uplink between the edge router and the query server as a representative constraint for real-time streaming. Based on this capacity, STAC inference stream operates at ~571 kbps, resulting in only 11.4\% bandwidth utilization. This optimized bandwidth usage enables the system to scale to multiple cameras while remaining responsive under constrained network conditions.
\vspace{-0.28cm}
\subsection{Qualitative analysis} \vspace{-0.2cm}Fig. 5(a) shows tiling and tile selection with RoIs capturing all key individuals, while Fig. 5(b) shows masked frame regions, reducing communication overhead.
\vspace{-0.2cm}
\begin{figure}[t]
    \centering
    \begin{minipage}[b]{0.49\linewidth}
        \centering
        \includegraphics[width=\textwidth, height=2.5cm]{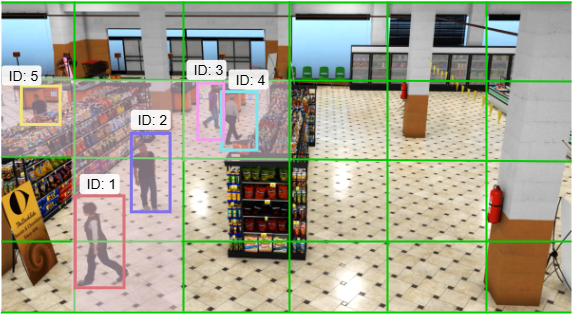}
        \subcaption{Tile selection with RoI}
        \label{fig:res1}
    \end{minipage}%  % The % here is CRUCIAL to eliminate space
    \begin{minipage}[b]{0.48\linewidth}
        \centering
        \includegraphics[width=\textwidth, height=2.5cm]{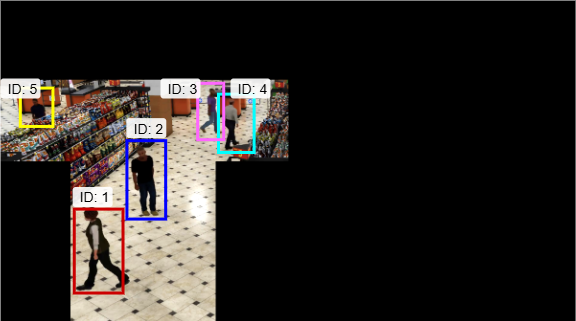}
        \subcaption{Masked output frame}
        \label{fig:res2}
    \end{minipage}
    \caption{\small Qualitative analysis of tiling and RoI based masking (Cam-2)}
    \label{fig:masking_results}
\end{figure}

\section{Conclusion and Future Work}  \vspace{-0.2cm}We present STAC, an object-wise spatio-temporal association that optimizes multi-camera, multi-target tracking, through system-level efficiencies and robust, fine-grained analytics. Future work includes outdoor camera testing and automated hyperparameters optimization.
\vspace{-0.45cm}
\section*{Acknowledgment} \vspace{-0.2cm} This work is supported by The Boeing Company under the University Master Research Agreement $\#2021-STU-PA-404$.
\vspace{-0.55cm}
\bibliographystyle{IEEEtran}
\vspace{-0.2cm}
\bibliography{stacmain}
\vspace{-0.3cm}

\end{document}